%% file: example_paper.tex
\documentclass{article}

\usepackage{microtype}
\usepackage{graphicx}
\usepackage{subcaption}
\usepackage{booktabs} 

\usepackage{hyperref}

\usepackage[preprint]{icml2026}

\usepackage{amsmath}
\usepackage{amssymb}
\usepackage{mathtools}
\usepackage{amsthm}

\usepackage[capitalize,noabbrev]{cleveref}

\theoremstyle{plain}

\theoremstyle{definition}

\theoremstyle{remark}

\usepackage[textsize=tiny]{todonotes}

\usepackage{pifont}
\usepackage{multirow}
\usepackage{colortbl}
\usepackage{arydshln}

\icmltitlerunning{VideoAesBench: Benchmarking the Video Aesthetics Perception Capabilities of LMMs}

\begin{document}

\twocolumn[
  \icmltitle{VideoAesBench: Benchmarking the Video Aesthetics Perception Capabilities of Large Multimodal Models}





  \begin{icmlauthorlist}
    \icmlauthor{Yunhao Li}{yyy}
    \icmlauthor{Sijing Wu}{yyy}
    \icmlauthor{Zhilin Gao}{yyy}
    \icmlauthor{Zicheng Zhang}{comp}
    \icmlauthor{Qi Jia}{comp}
    \icmlauthor{Huiyu Duan}{yyy} \\
    \icmlauthor{Xiongkuo Min}{yyy}
    \icmlauthor{Guangtao Zhai}{yyy}
  \end{icmlauthorlist}

  \icmlaffiliation{yyy}{Shanghai Jiao Tong University}
  \icmlaffiliation{comp}{Shanghai AI Laboratory}



  \vskip 0.3in
]

\printAffiliationsAndNotice{}  

\begin{abstract}

Large multimodal models (LMMs) have demonstrated outstanding capabilities in various visual perception tasks, which has in turn made the evaluation of LMMs significant. However, the capability of video aesthetic quality assessment, which is a fundamental ability for human, remains underexplored for LMMs. To address this, we introduce \textbf{VideoAesBench}, a comprehensive benchmark for evaluating LMMs’ understanding of video aesthetic quality. VideoAesBench has several significant characteristics: (1) Diverse content including 1,804 videos from multiple video sources including user-generated (UGC), AI-generated (AIGC), compressed, robotic-generated (RGC), and game videos. (2) Multiple question formats containing traditional single-choice questions, multi-choice questions, True or False questions, and a novel open-ended questions for video aesthetics description. (3) Holistic video aesthetics dimensions including visual form related questions from 5 aspects, visual style related questions from 4 aspects, and visual affectiveness questions from 3 aspects. Based on VideoAesBench, we benchmark 23 open-source and commercial large multimodal models. Our findings show that current LMMs only contain basic video aesthetics perception ability, their performance remains incomplete and imprecise. We hope our VideoAesBench can be served as a strong testbed and offer insights for explainable video aesthetics assessment. The data will be released on \url{https://github.com/michaelliyunhao/VideoAesBench}.

\end{abstract}

\section{Introduction}

\begin{figure*}[ht]
\centering
\includegraphics[width=\linewidth]{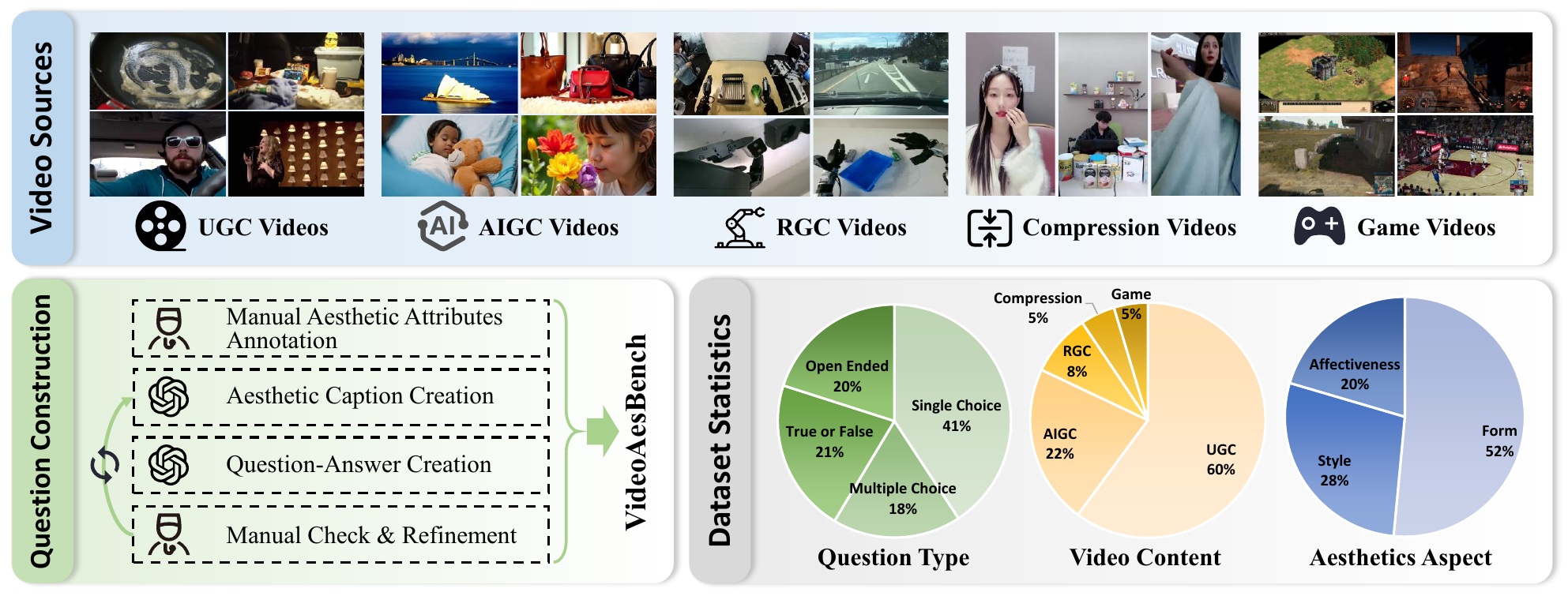}
\caption{Overview of our proposed \textbf{VideoAesBench}. The video are initially collected from five different scenarios including user-generated content (UGC) videos, AI-generated content (AIGC) videos, robot-generated content (RGC) videos, compression videos, and gaming videos. After obtaining the initial videos, we adopt a human-in-the-loop strategy to efficiently create high quality question-answer pairs for VideoAesBench. The concrete dataset distribution for question type, video content, and video aesthetics aspect are also depicted.}
\label{fig:framework}
\end{figure*}

\begin{table*}[t]
\centering
\renewcommand\arraystretch{1.1}

\resizebox{0.95\linewidth}{!}{
\begin{tabular}{l|c|c|c|c|c|c|c}
\bottomrule

\textbf{Benchmark}	& \textbf{Year}	& \textbf{Modality}		&	\textbf{Source Video} &	\textbf{Q\&A Pairs}	&	\textbf{Data Type}	&	\textbf{Question Type} &  \textbf{Dimensions}	\\
\hline

Q-Bench \cite{wu2023q}	&	2024	&	Image	&	2,990	&	2,990	&	UGC, Compress, AIGC 	&	SC, TF, OE &  4	\\

Q-Bench-Video \cite{zhang2025q}	& 2025 &		Video	&		1,800	&	2,378	&	UGC, AIGC, CG	&	SC, TF, OE  & 4	\\
\hdashline

AesBench \cite{huang2024aesbench}	& 2024  & Image	&	2,800	&	8,400	& UGC, AIGC, Art	&	SC, TF, OE & 8	\\

\rowcolor[gray]{.92}
\textbf{VideoAesBench (Ours)}	& 2026 & Video		&	1,641	&	1,804	&	UGC, AIGC, RGC, Compress, CG	&	SC, MC, TF, OE &  12	\\

\bottomrule
\end{tabular}
}

\caption{Summary and comparison of related benchmarks in quality assessment and aesthetics assessment. The abbreviations ``UGC", ``AIGC", ``CG", ``RGC", ``Compress", ``Art" represent user-generated content, AI-generated content, computer graphics content, robot-generated content, compression content, and artistic content, respectively.}
\centering
\label{tab:bench}
\vspace{-4mm}
\end{table*}

Video has been regarded as one of the most widely disseminated forms of content in the era of Internet with the help of long video and short video platforms, with numerous applications on social media, film and entertainment industries, as well as everyday life. Consequently, the understanding of video aesthetics has emerged as a significant problem in both academia and industry, which can evaluate and improve the user experiences of video content. With the rapid development of large multimodal models (LMMs), they have demonstrated outstanding capabilities in visual perception including both high-level semantic content understanding and low-level perceptual quality perception. However, their capabilities in understanding and evaluating video aesthetics still remain an unexplored problem.

To cope with this issue, a corresponding benchmark for LMMs is urgently needed. The evaluation benchmark for LMMs has been a hot topic in recent years, which aims at constructing a delicate exam for LMMs on various capabilities. Existing benchmarks \cite{liu2024mmbench} mainly focus evaluating the high-level semantic understanding abilities such as action recognition, spatial understanding. Few studies are conducted to evaluate the perceptual quality and aesthetic quality of generic images or videos. For instance, as summarized in Table \ref{tab:bench}, Q-Bench \cite{wu2023q} and Q-Bench-Video \cite{zhang2025q} evaluates the capabilities of LMMs in understanding the perceptual quality of images and videos, respectively. While AesBench \cite{huang2024aesbench} evaluates the the capabilities of LMMs in understanding the image aesthetics from 8 dimensions. However, a systematic LMM benchmark for video aesthetics has never been explored.

In the light of these facts, we introduce \textbf{VideoAesBench}, a carefully designed benchmark for video aesthetics perception. As demonstrated in Figure \ref{fig:framework}, VideoAesBench contains 1,804 video-question-answer triples from a wide range of video content including user-generated content (UGC), AI-generated content (AIGC), robot-generated content (RGC), compression, and gaming videos from 10 datasets. To comprehensively evaluate fine-grained aesthetics perception abilities of LMMs, all the questions are divided into three main aesthetic aspects including visual form, visual style, and visual affectiveness. Specifically, the visual form aspect includes visual composition, visual elements $\&$ structure, shot size, depth of field, and visual subject; the visual style aspect includes lighting, color, visual tone, and creativity; the visual affectiveness includes emotion, theme $\&$ communication, viewer interest. Meanwhile, we introduce diverse question types, including True or False, single choice, multiple choice, and open-ended questions.

Equipped with VideoAesBench, we comprehensively benchmark 18 open-source LMMs and 5 closed-source LMMs on video aesthetics perception and conclude several findings. First, Closed-source LMMs are generally better than open-source LMMs, except for Qwen3-VL. Second, multiple choice questions and open-ended questions are harder than single choice questions and True or False questions. Third, LMMs exhibit relatively unbalanced performance on various video aesthetic perspectives. Finally, different LMMs have their own unhandled types of videos. Overall, our main contributions are summarized as follows:

\begin{itemize} 
\item We propose \textbf{VideoAesBench}, the first comprehensive benchmark elaborately designed for evaluating the video aesthetics perception capabilities of LMMs, which contains diverse types of videos, multiple types of questions, 12 fine-grained video aesthetics dimensions.
\item Our evaluation benchmark consists of three main video aesthetics dimensions: \textit{Visual Form}, \textit{Visual Style}, and \textit{Visual Affectiveness}, offering a comprehensive discovery of aesthetics understanding capabilities. VideoAesBench also introduces a new question type, mlutiple choice question, for constructing challenging questions.
\item A comprehensive evaluation of both open-source and closed-source LMMs are conducted to measure their capabilities in understanding video aesthetics, revealing the strengths and weaknesses of each LMM and suggest valuable insights for explainable video aesthetics understanding.
\end{itemize}

\input{tabs/table_source}

\section{Related Work}

\begin{figure*}[h]
\centering
\includegraphics[width=0.95\linewidth]{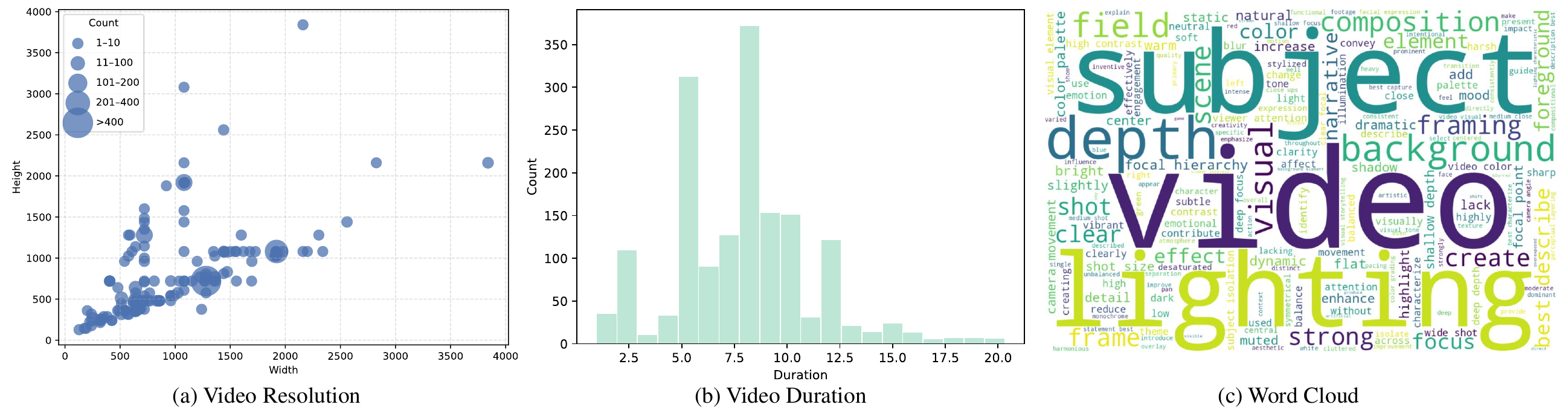}
\caption{Data statistics of VideoAesBench. (a): Video resolution distribution across video width and height in VideoAesBench. (b): Distribution of video duration. (c): The word cloud statistics of all questions in VideoAesBench.}
\label{fig:statistic}
\vspace{-1mm}
\end{figure*}

\subsection{Benchmarking Video LMMs}

Video understanding benchmarks \cite{AIBench} for video LMMs have been a hot topic in recent years, aiming at evaluating the capabilities of LMMs on various video-related tasks. Early benchmarks \cite{liu2024mmbench, chen2024we, zhang2025large} mainly focus on constructing a general exam for image LMMs with numerous tasks. Subsequent studies transform from image to video and construct various comprehensive video LMMs' benchmark including VideoMME \cite{fu2025video}, MVBench \cite{li2024mvbench} and so on. However, most of the studies pay attention on evluating high level semantic understanding capabilities. The perceptual quality and aesthetics quality perception capabilities are rarely explored. In these areas, Q-Bench \cite{wu2023q} and Q-Bench-Video \cite{zhang2025q} constructs holistic benchmarks designed for image and video perceptual quality assessment with open-ended questions and several video types. For visual aesthetics understanding, AesBench is the first holistic benchmark for image aesthetics perception, covering three key capabilities including aesthetic perception, aesthetic empathy, and aesthetic interpretation. In this paper, we introduce the first holistic benchmarks for video aesthetics understanding with 1,804 carefully designed questions covering 12 fine-grained aesthetics dimensions.

\subsection{Visual Aesthetics Assessment}

Visual aesthetics assessment is a long standing problem which predicts the average aesthetic evaluation of human users. Existing studies mainly focus on score-based image aesthetics assessment. Traditional methods utilize hand-crafted features \cite{dhar2011high, datta2006studying} to extract features of different visual aesthetic elements. With the development of neural network, numerous methods \cite{zhu2025progressively, li2024towards, he2023thinking, he2022rethinking, jin2024apddv2} utilize CNNs or transformers for better aesthetics score prediction. In recent years, based on LMMs, explainable image aesthetics assessment is becoming popular. AesBench \cite{huang2024aesbench} is first proposed to benchmark the image aesthetics understanding capabilities of LMMs, then AesExpert \cite{huang2024aesexpert} is proposed to construct a image aesthetics foundation model. As for video aesthetics assessment, DIVIDE-3K \cite{wu2023exploring} is first proposed with both perceptual quality scores and aesthetics scores. VADB \cite{qiao2025vadb} is recently proposed with fine-grained aesthetics scores, containing about 10K videos. In light of these facts, our paper is the first study exploring explainable video aesthetics assessment with large multimodal models.

\subsection{Visual Quality Assessment}

Apart from visual aesthetics assessment, visual quality assessment is a similar but different task which focuses on evaluating the perceptual quality of various visual modalities \cite{li2025dhqa, wei2026vtonqa, li2025exploring, wu2023ganhead, yang2025lmme3dhf, gao2025ges, li2025aghi,  gao2025multi}. For video quality assessment, early studies \cite{vu2011spatiotemporal, zheng2022completely} utilize hand-crafted features for regressing video quality scores. With the development of neural networks, convolution-based network \cite{you2019deep} and transformer-based network \cite{wu2023exploring} are designed for video quality assessment. In recent years, large multimodal model-based methods have gradually become state-of-the-art baseline methods for assessing general video quality \cite{ge2025lmm, zhou2026mi3s}, face video quality \cite{wu2025fvq}, and so on. Despite numerous efforts on utilizing LMMs for score regression, explainable video quality assessment has been rarely explored.

\begin{figure*}[ht]
\centering
\includegraphics[width=\linewidth]{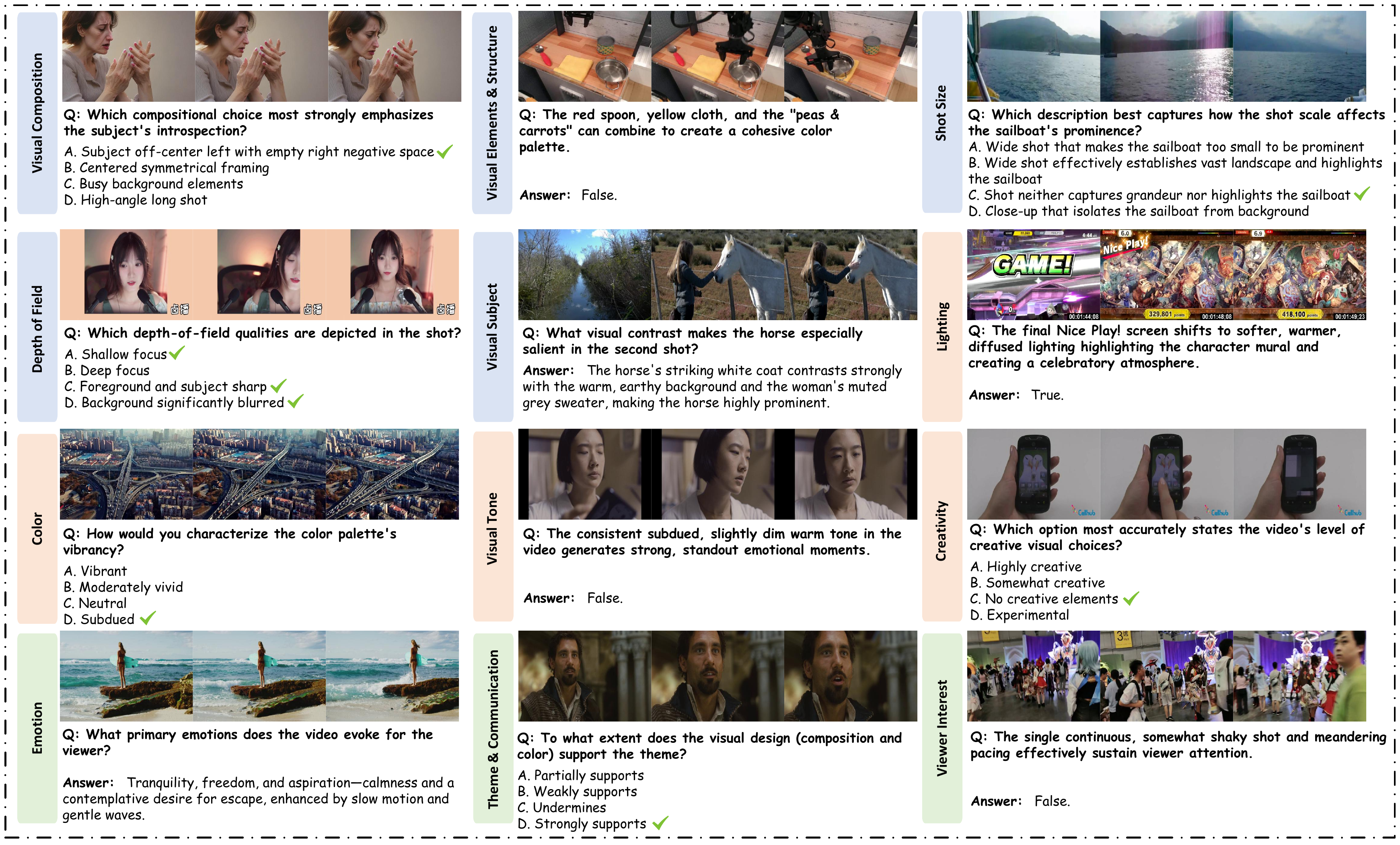}
\caption{Visualization examples from VideoAesBench with in terms of aesthetics dimension. In our benchmark, the question type contains single-choice question, multiple-choice question, True-or-Flase question, and open-ended question. The aesthetic dimension concretely contains visual composition, visual elements and structure, shot size, depth of field, visual subject, lighting, color, visual tone, creativity, emotion, theme and communication, and viewer interest.}
\label{fig:all_demos}
\vspace{-2mm}
\end{figure*}

\input{tabs/table1}

\input{tabs/table2}

\section{Benchmark Construction}

\subsection{Aesthetic Video Collection}

As summarized in Table \ref{tab:source}, to enable the diversity collected videos, We collect all raw videos from 10 different perceptual/aesthetic quality datasets covering 5 types of videos. The reason is that these datasets include various videos with both bad and good quality, which can cover a wide range of video content.

Concretely, we construct the raw video pool by combining 10 different datasets. The raw video pool includes (1) user-generated content (UGC) video datasets: LSVQ \cite{ying2021patch}, DIVIDE-3K \cite{wu2023towards}, VADB \cite{qiao2025vadb}, and FineVD \cite{duan2025finevq}, where LSVQ and FineVD are perceptual quality assessment datasets, DIVIDE-3K and VADB are aesthetics assessment datasets; (2) AI-generated content (AIGC) video datasets: Love \cite{wang2025love} and HVEval \cite{wu2025hveval}, where love is a general AI-generated video dataset and HVEval is an action-centric generated video dataset; (3) robot-generated content (RGC) video dataset: RGCD \cite{jin2025rgc}, which is collected from drone, wheel, and robot; (4) compression video dataset: TaoLive \cite{zhang2023md}, LIVE-Compress \cite{yu2021predicting}, where TaoLive is an compression dataset with live streaming videos collected from Taolive platform, LIVE-Compress contains videos with synthetic compression distortions; (5) gaming video dataset: LIVE-YT-Gaming \cite{yu2023subjective}, which is a live streaming gaming video dataset. After constructing the raw image pool, we conduct balanced sampling from each dataset considering its own scale.

\subsection{Question Design}

\noindent\textbf{True or False Questions.} We first construct True or False related questions as a fundamental questions which require large multimodal models to provide a binary judgments (True or False) considering the aesthetics description sentence of a video, focusing on evaluating the ability of aesthetics judgement for diverse videos. During the process of question construction, we ask the human annotators to ensure that the aesthetics description contains relevant information about the corresponding video, eliminating meaningless sentences. In addition, we specifically balance the ratio of True and False answers during the dataset collection process to avoid completely unbalanced ratios.

\noindent\textbf{Single Choice Questions.} Considering True or False questions sometimes are quite easy, we also introduce single choice questions for video aesthetics perception. Single choice question consists of a question sentence and four answer options represented by ethier phrases or sentences. The goal of LMMs is to select the most accurate answer from the set of options including distractors. This type of question can detailedly assess the capability of LMMs in understanding and distinguishing diverse aesthetics issues.

\noindent\textbf{Multiple Choice Questions.} Considering that single choice questions may not thoroughly assess the video aesthetics understanding ability of LMMs, we further include multiple choice questions which is considered the most difficult type of closed-ended questions. Multiple choice question specifically contains a question prompt and four answer options represented by ethier phrases or sentences. The goal of LMMs is to select all correct answer options (more than one) which accurately describe the aesthetics of a video. By constructing this type of questions, we can evaluate whether current LMMs can precieve all the aesthetic issues of general videos. Moreover, these questions can be treated as the most difficult parts of our benchmark, prompting future development of large multmodal models.

\noindent\textbf{Open-ended Questions.} All the three questions above are classified as closed-ended questions, however, human users may want large multimodal models to give a comprehensive aesthetics descriptions and suggestions for further development of a video. To precisely evaluate the capability of explainable video aesthetics understanding for LMMs, we finally introduce open-ended questions which do not restrict answers to predefined answer sets.

\subsection{Video Aesthetics Dimensions}

To conduct a comprehensive evaluation of current large multimodal models on video aesthetics understanding, we specifically design 3 basic aesthetic dimensions with 12 fine-grained aesthetic dimensions, which are summarized as follows:

\subsubsection{Visual Form.} Visual form evaluates how visual components are structurally organized within the frame, focusing on 5 sub-dimensions, including visual composition, Visual elements $\&$ structure, shot size, depth of field, and visual subject. \textbf{Visual composition} evaluates overall compositional organization, including balance, spatial rhythm, and focal hierarchy within the frame; \textbf{Visual elements $\&$ structure} evaluates how color, geometry, spatial arrangement, and lighting are structurally integrated to form a coherent visual design; \textbf{Shot size} evaluates the appropriateness of shot scale in establishing subject prominence and supporting narrative or expressive intent; \textbf{Depth of field} evaluates the use of depth of field as a visual mechanism for subject isolation and perceptual layering. \textbf{Visual subject} evaluates the subject's visual salience, recognizability, and expressive quality in terms of appearance, motion, and presence.

\subsubsection{Visual Style.} Visual style examines the aesthetic treatment of the image through 4 sub-dimensions including lighting, color, overall visual tone, and creativity, emphasizing how these elements shape mood, atmosphere, and stylistic identity. \textbf{Lighting} evaluates lighting quality and direction in shaping the subject and scene through effective illumination. \textbf{Color} evaluates color harmony and rendering, including naturalness or deliberate stylization in the visual presentation. \textbf{Visual tone} evaluates overall visual tone, such as brightness and color temperature, and its alignment with the intended mood or atmosphere. \textbf{Creativity} evaluates the originality of ideas and execution, focusing on inventive concepts or unconventional visual and narrative choices.

\subsubsection{Visual Affectiveness.} Visual affectiveness assesses the perceptual and emotional impact of the video, including emotional elicitation, thematic clarity, narrative communication, and the ability to capture and sustain viewer interest. We concretely divide visual affectiveness into three sub-dimensions including emotion, theme $\&$ communication, and viewer interest. \textbf{Emotion} evaluates the effectiveness of emotional elicitation, including the intensity and appropriateness of the emotions evoked. \textbf{Theme $\&$ communication} evaluates the clarity and coherence of the theme and the effectiveness of visual storytelling in conveying core ideas. \textbf{Viewer Interest} evaluates the capacity to capture and sustain viewer attention through visual appeal, pacing, and perceptual engagement.

\subsection{Question-Answer Pairs Construction}

Based on our collected videos, we specifically utilize a human-in-the-loop strategy for constructing question answer pairs. Concretely, we first recruit professional human annotators which covers a wide ages to annotate detailed descriptions of 12 fine-grained aesthetics dimensions according to our pre-defined definitions. Then, based on the annotated descriptions of all dimensions and the original video, we utilize Gemini-2.5 \cite{comanici2025gemini} to generate a complete aesthetics caption. With the obtained captions, we randomly assign a question type for each video and utilize GPT-5.2 to generate four different questions, covering various aesthetics dimensions. Finally, we recruit additional human annotators to manually check and refine the generated question and answer through a GUI software. Each question is checked by at least three human annotators. During the collecting process, if none of the four question-answer pairs are satisfactory, the videos are captioned again to ensure the correctness of each question-answer pair. After collection, VideoAesBench contains diverse video types and questions, which are shown in Figure \ref{fig:statistic} and \ref{fig:all_demos}.

\input{tabs/table3}

\begin{figure*}[ht]
\centering
\includegraphics[width=0.95\linewidth]{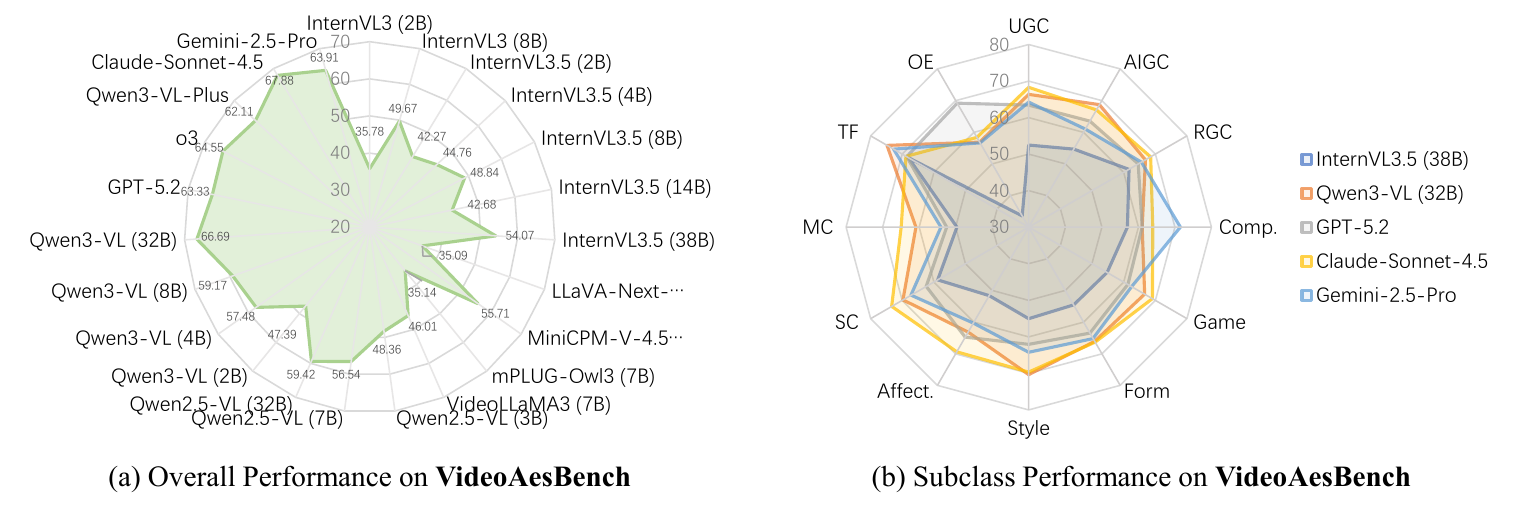}
\caption{Performance comparison of MLLMs on VideoAesBench. (a) Overall performance across all evaluated MLLMs. (b) Performance of five representative MLLMs across different subclasses, including video sources, aesthetic dimensions, and question types.}
\label{fig:radar}
\vspace{-4mm}
\end{figure*}

\section{VideoAesBench Experiments}

\subsection{Experimental Setup}

\noindent\textbf{Comparison LMMs.} To validate the perception ability of current large multimodal models (LMMs) on video aesthetics, we select 16 open-source models and 5 closed-source models on our VideoAesBench to conduct a comprehensive evaluation. Concretely, (1) Open-source LMMs contain: Qwen2.5-VL series \cite{bai2025qwen2}, Qwen3-VL series \cite{Qwen3-VL}, InternVL3 series \cite{zhu2025internvl3}, InternVL3.5 series \cite{wang2025internvl3}, VideoLLaMA3-7B \cite{zhang2025videollama}, mPLUG-Owl3-7B \cite{ye2024mplug}, MiniCPM-V-4.5 \cite{yu2025minicpm}, LLaVA-NeXT-Video-7B \cite{zhang2024llavanextvideo}; (2) Closed-source LMMs contain: Claude-Sonnet-4.5 \cite{claude}, Gemini-2.5-Pro \cite{comanici2025gemini}, GPT-5.2 \cite{}, Qwen-VL-Plus \cite{Qwen3-VL}, OpenAI O3 \cite{openai2025o3};

\noindent\textbf{Evaluation Metrics.} The questions in our VideoAesBench are divided into closed-ended (single-choice, multiple-choice, and True-or-False question types) and open-ended questions. For closed-ended questions, a answer is considered correct only if all and only the correct options are selected. As for open-ended questions, considering common rule-based evaluation method is not suitable, we employ GPT-5 to calculate a perception score by comparing the ground truth language answer and a LMM's answer. We specifically design a text prompt for evaluating the open-ended video aesthetics perception questions.

\subsection{Main Results}

The overall video aesthetics perception and fine-grained performance on \textbf{VideoAesBench} are summarized in Table \ref{tab:main} and Figure \ref{fig:radar}. Meanwhile, the LMMs' performance across various aesthetics dimensions are reported in Table \ref{tab:sub_dim}. The LMMs' performance across various video content are listed in Table \ref{tab:sub_imgtype_dim}. Through analyzing the results from all tables, several findings are discovered and shown as follows:

\textbf{1) General Performance. Closed-source LMMs are better than open-source LMMs, except for Qwen3-VL.} As observed from the results presented in Table \ref{tab:main}, we can find that all the large multimodal models with sizes larger than 4B significantly outperform the random guess baseline, demonstrating that current LMMs already contain the foundational ability to understand video aesthetics. Considering the open-source LMMs, we can observe that the Qwen3-VL (59.17$\%$), Qwen2.5-VL (56.54$\%$), and MiniCPM-V-4.5 (55.71$\%$) are the best three LMMs under the same model size (around 7B). For the InternVL series and Qwen-VL series, we can find that the performance consistently increases as the model version updates. As for the closed-source LMMs, with the help of the internal database, current LMMs generally perform better than open-source LMMs except for Qwen3-VL-32B. Among all models, Claude-Sonnet-4.5 and o3 are the best two LMMs. Considering all the LMMs, Claude-Sonnet-4.5 and Qwen3-VL-32B are the best two LMMs with quite good performance (67.88$\%$ and 66.69$\%$). Qwen3-VL is the only open-source model which is comparable against closed source LMMs. However, we can also observe that these two models still have room for improvement, inspiring for future development of video aesthetics understanding large multimodal models.

\textbf{2) Question type. Multiple choice questions and open-ended questions are harder than others.} As summarized in Table \ref{tab:main}, the difficulty of all types of question formats generally follow an intuitive trend: Multiple choice $>$ Open ended $>$ Single choice $>$ True or False. To be noticed, the best LMM, Claude-Sonnet-4.5, only achieves 64.91$\%$ for multiple choice questions, indicating that current
LMMs still struggle with diverse complex video aesthetics understanding tasks. Meanwhile, we can find that different LMMs are good at various question types: Qwen3-VL-32B performs best (74.81$\%$) on True or False questions, Claude-Sonnet-4.5 performs best on single choice questions (73.33$\%$) and multiple choice questions (64.91$\%$), and GPT-5.2 performs best (69.20$\%$) on open-ended questions.

\textbf{3) Aesthetics Dimension. LMMs exhibit relatively unbalanced performance on various video aesthetic perspectives.} As summarized in Table \ref{tab:sub_dim}, current LMMs demonstrate different abilities across various aesthetics dimensions. First, considering all the state-of-the-art LMMs, they mainly achieve stronger performance in understanding color, visual subject, and creativity. Surprisingly, LMMs achieve relatively high performance in understanding the creativity of videos, showing that abstract creative consciousness is easy for LMMs. Second, after inspecting each LMM, we can observe that the Qwen3-VL-32B model performs well in understanding visual element and structure, lighting, color, visual tone, creativity, and Theme $\&$ Communication; ChatGPT-o3 is good at understanding visual element and structure, depth of field; Claude-Sonnet-4.5 is good at understanding shot size, depth of field, visual subject, color, visual tone, emotion, and Theme $\&$ Communication; Gemini-2.5-Pro is good at understanding visual compostion and color.


\textbf{4) Video Sources. Different LMMs have their own unhandled types of videos.} As summarized in Table \ref{tab:main}, Table \ref{tab:sub_imgtype_dim}, we can observe that considering all state-of-the-art LMMs, current LMMs performs relatively balanced results on five kinds of videos including UGC, AIGC, RGC, compression, and gaming videos. It demonstates that, for video aesthetics understanding, current LMMs are able to handle diverse video types. Meanwhile, we can see that different LMMs have their own advantages: Qwen3-VL-32B specializes in understanding the aesthetics of AIGC videos;  Claude-Sonnet-4.5 is good at understanding  the aesthetics of UGC, RGC, and gaming videos; Gemini-2.5-Pro specializes in understanding the aesthetics of compression videos. Current LMMs have their own unbalanced video types, where a powerful LMM which is able to handle all types of videos is still needed.

\section{Conclusion}

In this paper, we introduce \textbf{VideoAesBench}, a comprehensive benchmark for evaluating LMMs’ understanding of video aesthetic quality. VideoAesBench has several significant characteristics: (1) Diverse content including 1,804 videos from multiple video sources including user-generated (UGC), AI-generated (AIGC), compressed, robotic-generated (RGC), and game videos. (2) Multiple question formats containing traditional single-choice questions, multi-choice questions, True or False questions, and a novel open-ended questions for video aesthetics description. (3) Holistic video aesthetics dimensions including visual form related questions from 5 aspects, visual style related questions from 4 aspects, and visual affectiveness questions from 3 aspects. Based on VideoAesBench, we benchmark 23 open-source and commercial large multimodal models. Our findings show that current LMMs only contain basic video aesthetics perception ability, their performance remains incomplete and imprecise. We hope our VideoAesBench can be served as a strong testbed and offer insights for explainable video aesthetics assessment.

\section{Broader Impact Statement}

VideoAesBench proposes a holistic benchmark for evaluating LMMs' video aesthetics perception capabilities for both real world videos and AI-generated videos. It is helpful for identifying the aesthetics issues of videos and providing improvement advise for optimizing both real captured videos and AI-generated videos.

\nocite{langley00}

\bibliography{example_paper}
\bibliographystyle{icml2026}

\newpage
\appendix
\onecolumn

\section{Appendix.}

\subsection{Open-ended Question Evaluation}

In this section, we specifically explain the instruction prompts which is used to guide GPT-5.2 for evaluating the quality of the answers of open-ended questions.

The prompt is ``You are a strict evaluator for open-ended question answering.\verb|\n| Given: - Question: \{question\}\verb|\n| - Reference Answer (ground truth): \{gt\}\verb|\n| - Model Response: \{pred\}\verb|\n| Your task is to evaluate the Model Response by comparing it ONLY against the Reference Answer. Do NOT use outside knowledge. \verb|\n| Scoring rules:\verb|\n| - Score 2:\verb|\n| The response is semantically equivalent to the reference answer. It is accurate, complete, and directly addresses the key points of the reference. Minor paraphrasing or wording differences are acceptable. \verb|\n| - Score 1:\verb|\n| The response is partially correct. It contains some correct information from the reference answer, but is missing key elements, is incomplete, or contains minor inaccuracies. \verb|\n|- Score 0:\verb|\n| The response is incorrect, irrelevant, contradicts the reference answer, or does
not meaningfully address the question. Also assign Score 0 if the response is empty, nonsensical, or not a valid answer. \verb|\n| Output format (STRICT):
$<$0$|$1$|$2$>$ Do not provide explanations or any additional text."

\subsection{GUI demo}

\begin{figure*}[ht]
\centering
\includegraphics[width=0.95\linewidth]{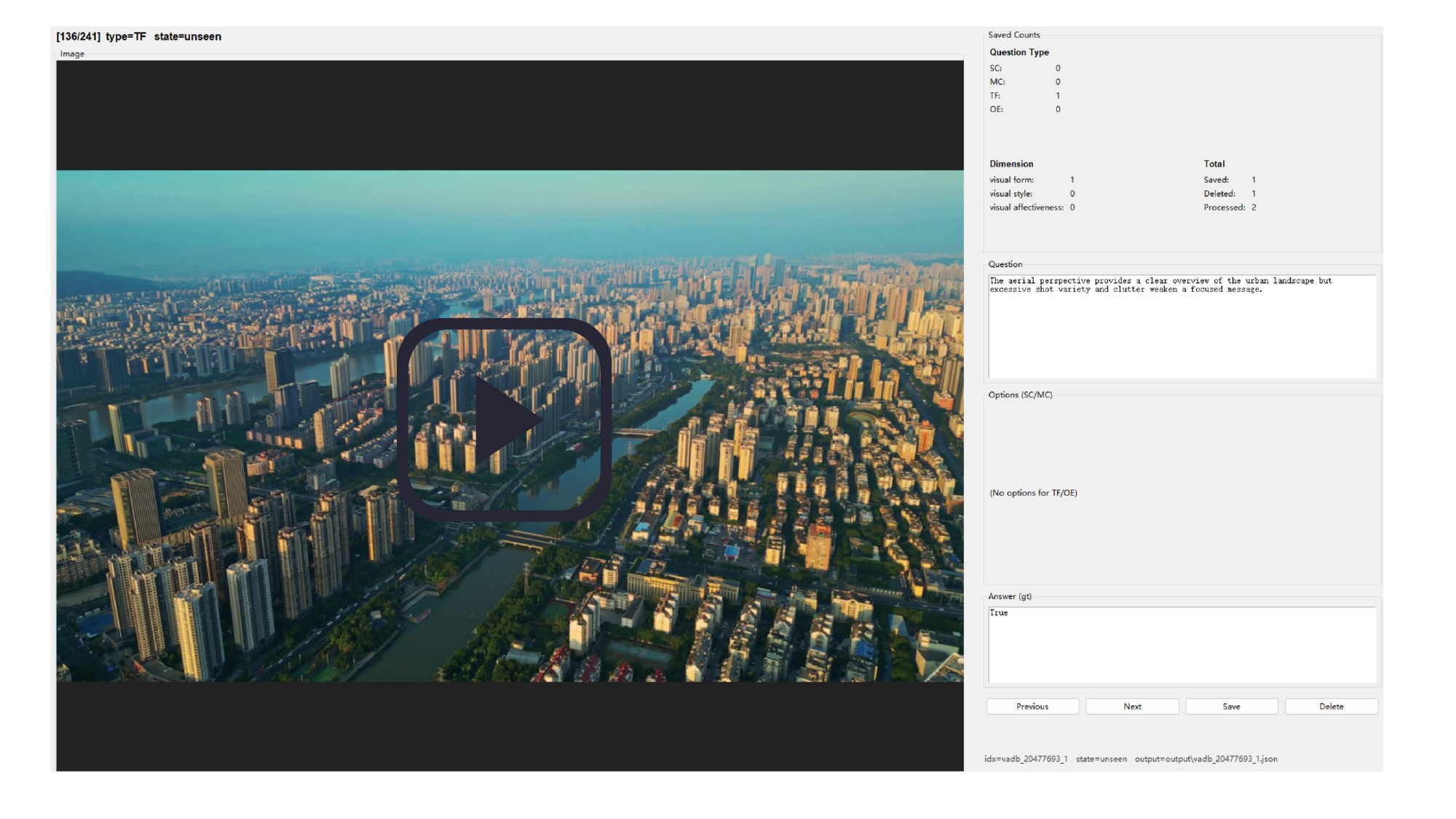}
\caption{GUI Figure.}
\label{fig:gui}
\end{figure*}

We also provide the GUI figure for human annotators during the collection of our VideAesBench. The figure is shown in Figure \ref{fig:gui}.

\end{document}

%% file: tabs/table_source.tex
\begin{table*}[!t]
    \centering
    \renewcommand\arraystretch{1.1}
    \setlength{\tabcolsep}{13pt}
    
   \resizebox{0.95\linewidth}{!}{\begin{tabular}{l|l|c|c|c|c}
    \bottomrule
    \textbf{Video Type} & \textbf{Video Source Dataset} & \textbf{Annotation} & \textbf{Dataset Information} & \textbf{Full Dataset Size}  & \textbf{Sampled Size} \\
   \hline
    \multirow{4}{*}{UGC (1085)} & LSVQ~\cite{ying2021patch} & Perceptual Score & Numerous Internet UGC Videos & 39K  & 271\\
    & DIVIDE-3K~\cite{wu2023towards} & Perceptual $\&$ Aesthetic Score & Mainly YouTube Videos & 3,590 & 152 \\
    & VADB~\cite{qiao2025vadb} & Aesthetic Score & Documentaries,
    Films, TV Dramas, etc & 10.5K  & 404  \\
    & FineVD~\cite{duan2025finevq} & Perceptual Score & Common $\&$ Short-form UGC Videos & 6,104 & 258 \\ \hdashline
    \multirow{2}{*}{AIGC (395)} & Love~\cite{wang2025love} & Perceptual Score & Generated from 30 Text-to-video Models & 58.5K & 279\\
    & HVEval~\cite{wu2025hveval} & Perceptual Score & Generated from 24 Text-to-video Models & 20K &116 \\
    \hdashline
    RGC (154) & RGCD~\cite{jin2025rgc} & Perceptual Score  & Wheel $\&$ Drone $\&$ Robot Videos & 2,100 & 154 \\
    \hdashline
    \multirow{2}{*}{Compression (86)} & TaoLive~\cite{zhang2023md}  &  Perceptual Score & Compressed Taolive Platform Videos &  3,762 & 82 \\
    &  LIVE-Compress~\cite{yu2021predicting} & Perceptual Score & Compressed Camera-captured Videos & 275 & 4 \\
    \hdashline
    Game (84) & LIVE-YT-Gaming~\citep{yu2023subjective} & Perceptual Score &  Real Gaming Videos on YouTube & 600 & 84 \\
    \bottomrule
\end{tabular}}
\caption{Overview of the all 10 video candidate datasets in the \textbf{VideoAesBench}. Multiple video content types are collected in our dataset, including \textbf{user-generated content (UGC) videos, AI-generated content (AIGC) videos, robot-generated content (RGC) videos, compression videos, and gaming videos}.}
\label{tab:source}
\vspace{-2mm}
\end{table*}

%% file: tabs/table1.tex
\begin{table*}[t]
\centering
\renewcommand\arraystretch{1.1}
\small
\resizebox{\textwidth}{!}{
\begin{tabular}{l ccccc ccc cccc c}
\toprule
\multirow{2}{*}{\textbf{Model}} & \multicolumn{5}{c}{\textbf{Video Sources}} & \multicolumn{3}{c}{\textbf{Aesthetic Dimensions}} & \multicolumn{4}{c}{\textbf{Question Types}} & \multirow{2}{*}{\textbf{Overall}\,$\uparrow$} \\
\cmidrule(lr){2-6} \cmidrule(lr){7-9} \cmidrule(lr){10-13}

& UGC & AIGC & RGC & Comp. & Game & Form & Style & Affect. & SC & MC & TF & OE \\

\midrule

\rowcolor[gray]{.92}
\multicolumn{14}{c}{\textit{Baseline}} \\  
\hdashline

Random guess & 33.26 & 34.51 & 35.11 & 31.03 & 33.80 & 33.85 & 33.44 & 33.11 & 25.00 & - & 50.00 & - & 33.59 \\

\hdashline
\rowcolor[gray]{.92}
\multicolumn{14}{c}{\textit{Open-source MLLMs}} \\  
\hdashline

InternVL3 (2B) \cite{zhu2025internvl3} & 34.19 & 41.14 & 34.42 & 29.65 & 39.88 & 35.74 & 36.07 & 35.50 & 37.01 & 21.43 & 57.66 & 22.79 & 35.78 \\
InternVL3 (8B) \cite{zhu2025internvl3} & 48.85 & 48.86 & 54.22 & 51.16 & 54.17 & 49.46 & 51.68 & 47.43 & 56.87 & 39.44 & 65.45 & 27.35 & 49.67 \\
InternVL3.5 (2B) \cite{wang2025internvl3} & 41.29 & 43.04 & 44.81 & 39.53 & 49.40 & 41.17 & 42.98 & 44.04 & 45.85 & 32.30 & 55.58 & 29.70 & 42.27 \\
InternVL3.5 (4B) \cite{wang2025internvl3} & 43.78 & 46.58 & 44.48 & 45.35 & 48.81 & 42.25 & 49.70 & 44.31 & 53.33 & 28.88 & 60.00 & 25.28 & 44.76 \\
InternVL3.5 (8B) \cite{wang2025internvl3} & 47.93 & 51.90 & 46.75 & 38.95 & 60.12 & 47.79 & 51.98 & 47.15 & 54.56 & 39.44 & 63.90 & 29.56 & 48.84 \\
InternVL3.5 (14B) \cite{wang2025internvl3} & 42.49 & 40.63 & 43.83 & 50.00 & 45.24 & 42.84 & 44.27 & 40.11 & 37.96 & 43.17 & 61.56 & 31.77 & 42.68 \\
InternVL3.5 (38B) \cite{wang2025internvl3} & 52.49 & 54.68 & 61.69 & 56.98 & 54.76 & 54.52 & 55.04 & 51.63 & 58.78 & 49.69 & 68.05 & 33.56 & 54.07 \\
LLaVA-Next-Video (7B) \cite{zhang2024video}  & 33.82 & 37.22 & 36.04 & 32.56 & 42.26 & 33.58 & 36.26 & 37.26 & 31.97 & 18.94 & 57.40 & 32.04 & 35.09 \\
MiniCPM-V-4.5 (7B) \cite{yu2025minicpm} & 54.10 & 58.73 & 53.90 & 59.30 & 61.90 & 54.09 & 59.09 & 55.15 & 62.31 & 50.93 & 64.42 & 37.29 & 55.71 \\
mPLUG-Owl3 (7B) \cite{ye2024mplug} & 33.78 & 35.70 & 41.23 & 30.81 & 43.45 & 33.91 & 37.45 & 35.09 & 24.63 & 27.02 & 66.23 & 30.66 & 35.14 \\
VideoLLaMA3 (7B) \cite{zhang2025videollama} & 42.81 & 52.66 & 52.27 & 33.14 & 57.74 & 44.83 & 45.26 & 50.00 & 44.63 & 31.99 & 70.91 & 34.81 & 46.01 \\
Qwen2.5-VL (3B) \cite{bai2025qwen2} & 46.45 & 53.29 & 42.86 & 52.33 & 55.95 & 46.66 & 49.31 & 51.36 & 53.33 & 36.34 & 63.12 & 33.29 & 48.36 \\
Qwen2.5-VL (7B) \cite{bai2025qwen2} & 54.19 & 62.91 & 58.44 & 49.42 & 60.71 & 56.08 & 55.34 & 59.35 & 61.77 & 44.10 & 71.17 & 41.44 & 56.54 \\
Qwen2.5-VL (32B) \cite{bai2025qwen2} & 58.76 & 60.89 & 62.99 & 58.14 & 55.95 & 57.53 & 63.93 & 57.99 & 64.49 & 49.38 & 68.31 & 48.62 & 59.42 \\
Qwen3-VL (2B) \cite{Qwen3-VL} & 45.71 & 50.51 & 48.05 & 42.44 & 58.33 & 47.69 & 48.22 & 45.53 & 46.53 & 40.99 & 62.60 & 38.67 & 47.39 \\
Qwen3-VL (4B) \cite{Qwen3-VL} & 55.94 & 59.75 & 61.04 & 53.49 & 64.29 & 56.03 & 60.67 & 56.78 & 62.59 & 50.00 & 69.09 & 41.44 & 57.48 \\
Qwen3-VL (8B) \cite{Qwen3-VL} & 57.97 & 61.39 & 60.06 & 52.91 & \underline{69.05} & 58.77 & 59.88 & 59.21 & 61.90 & 52.17 & 70.65 & 47.65 & 59.17 \\
Qwen3-VL (32B) \cite{Qwen3-VL} & 66.36 & \textbf{68.73} & 66.88 & 61.05 & 66.67 & 66.09 & \textbf{70.36} & 63.14 & \underline{69.80} & \underline{60.87} & \textbf{74.81} & 56.91 & \underline{66.69} \\

\hdashline
\rowcolor[gray]{.92}
\multicolumn{14}{c}{\textit{Closed-source MLLMs}} \\  
\hdashline

GPT-5.2 \cite{singh2025openai} & 63.41 & 63.54 & 64.61 & 61.05 & 61.31 & 63.46 & 62.06 & \underline{64.77} & 62.45 & 52.48 & 68.57 & \textbf{69.20} & 63.33 \\
o3 \cite{openai2025o3} & \underline{66.82} & 58.99 & \underline{67.86} & \underline{66.28} & 53.57 & \underline{66.31} & 65.61 & 58.67 & 66.80 & 59.63 & 66.75 & \underline{62.02} & 64.55 \\
Qwen3-VL-Plus \cite{Qwen3-VL} & 61.66 & 62.03 & 64.61 & 59.30 & 66.67 & 61.19 & 64.72 & 60.84 & 61.22 & 52.80 & \underline{73.51} & 60.08 & 62.11 \\
Claude-Sonnet-4.5 \cite{claude} & \textbf{68.29} & \underline{67.09} & \textbf{68.51} & 63.95 & \textbf{69.05} & \textbf{66.31} & \underline{69.66} & \textbf{69.38} & \textbf{73.33} & \textbf{64.91} & 68.83 & 58.43 & \textbf{67.88} \\
Gemini-2.5-Pro \cite{comanici2025gemini} & 64.24 & 61.01 & 65.58 & \textbf{71.51} & 62.50 & 65.18 & 64.23 & 60.30 & 67.21 & 54.04 & 72.73 & 56.63 & 63.91 \\

\bottomrule
\end{tabular}
}
\caption{Experimental results on VideoAesBench for video aesthetic quality perception capability of LMMs. The abbreviations ``Form", ``Style", and ``Affect." denote three aesthetic perspectives: visual form, visual style, and visual affectiveness, respectively. ``Comp." denotes compression videos. The best and runner-up performances are bold and underlined, respectively.}
\vspace{-4mm}
\label{tab:main}
\end{table*}

%% file: tabs/table2.tex
\begin{table*}[t]
\centering
\renewcommand\arraystretch{1.1}
\small
\resizebox{\textwidth}{!}{
\begin{tabular}{l ccccc ccc cccc}
\toprule
\multirow{2}{*}{\textbf{Model}} & \multicolumn{5}{c}{\textbf{Visual Form}} & \multicolumn{4}{c}{\textbf{Visual Style}} & \multicolumn{3}{c}{\textbf{Visual Affectiveness}} \\
\cmidrule(lr){2-6} \cmidrule(lr){7-10} \cmidrule(lr){11-13}

& Compos. & Element & Shot & Depth & Subject & Lighting & Color & Tone & Creativity & Emotion & Theme & Interest \\

\midrule

\rowcolor[gray]{.92}
\multicolumn{13}{c}{\textit{Baseline}} \\  
\hdashline

Random guess & 33.26 & 35.10 & 35.33 & 33.07 & 34.62 & 33.64 & 31.33 & 36.11 & 34.29 & 33.20 & 30.95 & 34.47  \\

\hdashline
\rowcolor[gray]{.92}
\multicolumn{13}{c}{\textit{Open-source MLLMs}} \\  
\hdashline

InternVL3 (2B) \cite{zhu2025internvl3} & 36.44 & 34.21 & 41.60 & 33.84 & 31.13 & 36.58 & 34.56 & 41.53 & 34.30 & 43.22 & 37.39 & 27.50 \\
InternVL3 (8B) \cite{zhu2025internvl3} & 48.87 & 48.83 & 50.42 & 51.29 & 45.28 & 51.32 & 54.78 & 47.46 & 50.83 & 52.54 & 50.90 & 40.36 \\
InternVL3.5 (2B) \cite{wang2025internvl3} & 38.98 & 42.98 & 45.80 & 40.30 & 43.40 & 39.21 & 45.59 & 44.07 & 45.45 & 44.07 & 40.54 & 46.79 \\
InternVL3.5 (4B) \cite{wang2025internvl3} & 39.41 & 47.08 & 49.16 & 39.44 & 42.45 & 46.84 & 54.78 & 51.69 & 47.52 & 48.31 & 47.30 & 38.57 \\
InternVL3.5 (8B) \cite{wang2025internvl3} & 47.46 & 47.08 & 47.90 & 48.06 & 50.94 & 55.53 & 49.26 & 47.46 & 51.65 & 53.39 & 50.90 & 38.93 \\
InternVL3.5 (14B) \cite{wang2025internvl3} & 38.70 & 43.27 & 49.58 & 46.55 & 37.74 & 41.84 & 49.26 & 50.85 & 39.26 & 38.98 & 43.24 & 38.57 \\
InternVL3.5 (38B) \cite{wang2025internvl3} & 54.52 & 52.92 & 55.04 & 54.53 & 58.49 & 51.84 & 55.88 & 61.86 & 55.79 & 55.51 & 51.35 & 48.57 \\
LLaVA-Next-Video (7B) \cite{zhang2024llavanextvideo} & 32.20 & 31.58 & 36.55 & 34.48 & 38.68 & 43.68 & 26.84 & 45.76 & 30.58 & 42.80 & 38.29 & 31.79 \\
MiniCPM-V-4.5 (7B) \cite{yu2025minicpm} & 53.95 & 50.00 & 57.14 & 56.03 & 52.83 & 57.37 & 62.87 & 56.78 & 58.68 & 62.29 & 60.81 & 44.64 \\
mPLUG-Owl3 (7B) \cite{ye2024mplug} & 32.34 & 34.80 & 44.12 & 30.60 & 33.02 & 41.58 & 34.93 & 43.22 & 30.99 & 47.88 & 28.38 & 29.64 \\
VideoLLaMA3 (7B) \cite{zhang2025videollama} & 43.93 & 50.00 & 51.26 & 42.24 & 31.13 & 43.16 & 42.28 & 54.24 & 47.52 & 52.12 & 48.20 & 49.64 \\
Qwen2.5-VL (3B) \cite{bai2025qwen2} & 47.46 & 43.27 & 53.36 & 43.53 & 50.94 & 49.21 & 54.41 & 49.15 & 43.80 & 56.78 & 47.75 & 49.64 \\
Qwen2.5-VL (7B) \cite{bai2025qwen2} & 58.47 & 54.68 & 60.92 & 48.71 & 66.04 & 53.68 & 59.56 & 47.46 & 57.02 & 59.75 & 59.46 & 58.93 \\
Qwen2.5-VL (32B) \cite{bai2025qwen2} & 53.53 & 62.57 & 61.76 & 58.41 & 54.72 & 59.74 & 68.75 & 62.71 & 65.70 & 66.95 & 53.15 & 54.29 \\
Qwen3-VL (2B) \cite{Qwen3-VL} & 46.89 & 47.08 & 53.36 & 45.26 & 52.83 & 53.68 & 47.06 & 49.15 & 40.50 & 51.69 & 45.50 & 40.36 \\
Qwen3-VL (4B) \cite{Qwen3-VL} & 53.53 & 56.14 & 61.34 & 56.47 & 58.49 & 62.11 & 61.76 & 57.63 & 58.68 & 64.83 & 59.46 & 47.86 \\
Qwen3-VL (8B) \cite{Qwen3-VL} & 58.19 & 62.28 & 57.98 & 58.62 & 53.77 & 57.89 & 64.71 & 61.86 & 56.61 & 60.17 & 60.81 & 57.14 \\
Qwen3-VL (32B) \cite{Qwen3-VL} & \underline{65.25} & \textbf{68.71} & \underline{65.55} & 66.38 & 63.21 & \textbf{71.05} & \underline{72.43} & \underline{63.56} & \textbf{70.25} & 63.56 & \textbf{68.02} & 58.93 \\

\hdashline
\rowcolor[gray]{.92}
\multicolumn{13}{c}{\textit{Commercial MLLMs}} \\  
\hdashline

GPT-5.2 \cite{singh2025openai} & 61.16 & 67.84 & 63.87 & 61.64 & \underline{71.70} & 61.32 & 57.72 & 59.32 & \underline{69.42} & \underline{67.37} & 64.86 & \underline{62.50} \\
o3 \cite{singh2025openai} & 64.55 & \underline{68.42} & 63.45 & \textbf{68.53} & 67.92 & 63.95 & 66.18 & 66.10 & 67.36 & 58.47 & 55.86 & 61.07 \\
Qwen3-VL-Plus \cite{Qwen3-VL} & 61.72 & 64.91 & 63.87 & 55.82 & 63.21 & 65.79 & 70.22 & 59.32 & 59.50 & 64.83 & 60.81 & 57.50 \\
Claude-Sonnet-4.5 \cite{claude} & 63.98 & 66.08 & \textbf{68.49} & \underline{67.67} & \textbf{71.70} & 65.00 & \textbf{77.21} & \textbf{69.49} & 68.60 & \textbf{71.61} & \underline{65.32} & \textbf{70.71} \\
Gemini-2.5-Pro \cite{comanici2025gemini} & \textbf{67.94} & 64.91 & 57.56 & 64.66 & 66.98 & \underline{66.05} & 67.65 & 59.32 & 59.92 & 64.41 & 61.71 & 55.71 \\

\bottomrule
\end{tabular}
}

\caption{Experimental results on VideoAesBench for video aesthetics quality perception capabilities of LMMs across different aesthetics dimensions. ``Compos.", ``Element", ``Shot", ``Depth", and ``Subject" denote visual composition, visual elements and structure, shot size, depth of field, and visual subject, respectively. The best and runner-up performances are bold and underlined, respectively.}
\label{tab:sub_dim}
\vspace{-6mm}
\end{table*}

%% file: tabs/table3.tex
\begin{table*}[ht]
\centering
\renewcommand\arraystretch{1.1}
\small
\resizebox{\textwidth}{!}{
\begin{tabular}{l ccc ccc ccc ccc ccc}
\toprule
\multirow{2}{*}{\textbf{Model}} & \multicolumn{3}{c}{\textbf{UGC}} & \multicolumn{3}{c}{\textbf{AIGC}} & \multicolumn{3}{c}{\textbf{RGC}} & \multicolumn{3}{c}{\textbf{Compression}} & \multicolumn{3}{c}{\textbf{Game}} \\
\cmidrule(lr){2-4} \cmidrule(lr){5-7} \cmidrule(lr){8-10} \cmidrule(lr){11-13} \cmidrule(lr){14-16}
& Form & Style & Affect. & Form & Style & Affect. & Form & Style & Affect. & Form & Style & Affect. & Form & Style & Affect. \\
\midrule

\rowcolor[gray]{.92}
\multicolumn{16}{c}{\textit{Baseline}} \\  
\hdashline

Random guess & 33.18 & 33.05 & 33.86 & 36.26 & 34.27 & 32.09 & 34.46 & 37.50 & 31.25 & 33.33 & 25.00 & 30.00 & 33.33 & 31.25 & 37.50  \\

\hdashline
\rowcolor[gray]{.92}
\multicolumn{16}{c}{\textit{Open-source MLLMs}} \\  
\hdashline

InternVL3 (2B) \cite{zhu2025internvl3} & 35.34 & 32.65 & 32.99 & 44.74 & 38.56 & 38.21 & 25.00 & 42.86 & 37.04 & 21.57 & 33.33 & 47.50 & 38.04 & 52.63 & 31.58 \\
InternVL3 (8B) \cite{zhu2025internvl3} & 48.24 & 51.89 & 46.19 & 50.29 & 48.31 & 47.17 & 51.56 & 53.17 & 62.96 & 52.94 & 46.67 & 50.00 & 55.43 & 68.42 & 36.84 \\
InternVL3.5 (2B) \cite{wang2025internvl3} & 40.45 & 41.41 & 43.65 & 44.44 & 41.53 & 42.45 & 40.62 & 48.41 & 46.30 & 35.29 & 43.33 & 47.50 & 45.65 & 57.89 & 50.00 \\
InternVL3.5 (4B) \cite{wang2025internvl3} & 42.13 & 50.00 & 39.59 & 44.44 & 49.15 & 47.17 & 35.94 & 44.44 & 64.81 & 37.25 & 50.00 & 62.50 & 50.00 & 65.79 & 28.95 \\
InternVL3.5 (8B) \cite{wang2025internvl3} & 47.99 & 50.34 & 44.16 & 49.71 & 53.39 & 53.77 & 44.53 & 47.62 & 50.00 & 33.33 & 70.00 & 30.00 & 58.70 & 68.42 & 55.26 \\
InternVL3.5 (14B) \cite{wang2025internvl3} & 42.04 & 45.02 & 40.10 & 42.11 & 41.53 & 37.26 & 45.31 & 46.03 & 35.19 & 48.04 & 56.67 & 50.00 & 46.74 & 34.21 & 52.63 \\
InternVL3.5 (38B) \cite{wang2025internvl3} & 53.85 & 51.72 & 49.49 & 56.73 & 53.39 & 52.83 & 58.59 & 65.08 & 61.11 & 51.96 & 70.00 & 60.00 & 52.17 & 71.05 & 44.74 \\
LLaVA-Next-Video (7B) \cite{zhang2024llavanextvideo} & 32.08 & 34.36 & 38.32 & 39.18 & 34.75 & 36.79 & 33.59 & 43.65 & 24.07 & 27.45 & 30.00 & 47.50 & 39.13 & 55.26 & 36.84 \\
MiniCPM-V-4.5 (7B) \cite{yu2025minicpm} & 53.10 & 56.36 & 53.81 & 57.89 & 61.86 & 56.60 & 48.44 & 58.73 & 55.56 & 50.00 & 70.00 & 75.00 & 65.22 & 76.32 & 39.47 \\
mPLUG-Owl3 (7B) \cite{ye2024mplug} & 32.66 & 35.40 & 34.77 & 35.38 & 36.44 & 35.38 & 35.94 & 46.83 & 40.74 & 30.39 & 30.00 & 32.50 & 45.65 & 50.00 & 31.58 \\ 
VideoLLaMA3 (7B) \cite{zhang2025videollama} & 43.22 & 39.52 & 46.45 & 51.17 & 55.08 & 52.36 & 42.19 & 54.76 & 70.37 & 31.37 & 30.00 & 40.00 & 60.87 & 52.63 & 55.26 \\
Qwen2.5-VL (3B) \cite{bai2025qwen2} & 44.05 & 48.45 & 50.76 & 56.43 & 50.00 & 51.89 & 36.72 & 49.21 & 42.59 & 45.10 & 56.67 & 67.50 & 59.78 & 52.63 & 50.00 \\
Qwen2.5-VL (7B) \cite{bai2025qwen2} & 52.85 & 53.61 & 59.14 & 66.96 & 57.63 & \underline{62.26} & 64.06 & 54.76 & 53.70 & 36.27 & 56.67 & \underline{77.50} & 68.48 & 68.42 & 34.21 \\
Qwen2.5-VL (32B) \cite{bai2025qwen2} & 58.71 & 61.17 & 55.33 & 57.31 & 66.95 & 59.91 & 59.38 & 66.67 & 62.96 & 47.06 & 76.67 & 72.50 & 52.17 & 68.42 & 52.63 \\
Qwen3-VL (2B) \cite{Qwen3-VL} & 45.39 & 46.05 & 46.19 & 56.14 & 49.15 & 42.92 & 42.19 & 53.97 & 48.15 & 43.14 & 36.67 & 45.00 & 58.70 & 65.79 & 50.00 \\
Qwen3-VL (4B) \cite{Qwen3-VL} & 55.70 & 56.87 & 55.33 & 59.36 & 62.71 & 57.08 & 50.78 & \underline{68.25} & 68.52 & 46.08 & 70.00 & 60.00 & 66.30 & 73.68 & 50.00 \\
Qwen3-VL (8B) \cite{Qwen3-VL} & 58.38 & 57.90 & 56.85 & 61.99 & 61.86 & 59.91 & 57.81 & 59.52 & 66.67 & 42.16 & 70.00 & 67.50 & \textbf{71.74} & 71.05 & \underline{60.53} \\
Qwen3-VL (32B) \cite{Qwen3-VL} & 65.41 & \underline{70.10} & 63.71 & \textbf{71.35} & \textbf{72.03} & 60.85 & \underline{65.62} & 65.87 & 72.22 & 53.92 & 70.00 & 72.50 & \underline{69.57} & 78.95 & 47.37 \\

\hdashline
\rowcolor[gray]{.92}
\multicolumn{16}{c}{\textit{Commercial MLLMs}} \\  
\hdashline

GPT-5.2 \cite{singh2025openai} & 62.90 & 63.06 & \underline{65.48} & 68.13 & 59.32 & 60.85 & 63.28 & 63.49 & 70.37 & 56.86 & 56.67 & 75.00 & 60.87 & 63.16 & 60.53 \\ 
o3 \cite{singh2025openai} & \textbf{68.68} & 66.15 & 62.18 & 64.33 & 61.44 & 47.64 & \textbf{69.53} & 61.90 & \underline{77.78} & \underline{61.76} & \underline{76.67} & 70.00 & 43.48 & \textbf{86.84} & 44.74 \\
Qwen3-VL-Plus \cite{Qwen3-VL} & 61.06 & 62.54 & 62.18 & 61.70 & 66.10 & 58.02 & 65.62 & \textbf{69.84} & 50.00 & 49.02 & 70.00 & \textbf{77.50} & 68.48 & 68.42 & 60.53 \\
Claude-Sonnet-4.5 \cite{claude} & 66.25 & \textbf{71.13} & \textbf{70.30} & \underline{68.71} & \underline{67.37} & \textbf{64.15} & 66.41 & 64.29 & \textbf{83.33} & 57.84 & 70.00 & 75.00 & 67.39 & 78.95 & \textbf{63.16} \\
Gemini-2.5-Pro \cite{comanici2025gemini} & \underline{66.58} & 62.54 & 59.64 & 63.16 & 61.44 & 57.08 & 60.16 & 67.46 & 74.07 & \textbf{68.63} & \textbf{83.33} & 70.00 & 57.61 & \underline{81.58} & 55.26 \\

\bottomrule
\end{tabular}
}
\caption{Experimental results on VideoAesBench for video aesthetic quality perception capabilities of LMMs across different video sources. The best and runner-up performances are bold and underlined, respectively. ``Form", ``Style", ``Affect" denote visual form, visual style, and visual affectiveness aspects.}
\label{tab:sub_imgtype_dim}
\vspace{-4mm}
\end{table*}